\pdfoutput=1

\documentclass[11pt]{article}

\usepackage[]{emnlp2021}

\usepackage{times}
\usepackage{latexsym}

\usepackage[T1]{fontenc}

\usepackage[utf8]{inputenc}

\usepackage{microtype}

\newcommand{\spert}{SpERT\xspace}

\newcommand{\scibert}{SciBERT\xspace}
\newcommand{\scierc}{SciERC\xspace}
\newcommand{\sciclaim}{SciClaim\xspace}

\newcommand{\figref}[1]{Figure~\ref{fig:#1}}

\newcommand{\tabref}[1]{Table~\ref{tab:#1}}

\newcommand{\textq}[1]{``\textit{#1}''}

\newcommand{\hide}[1]{}

\usepackage{booktabs}
\usepackage{makecell}
\usepackage{array,multirow,graphicx}
\usepackage{xspace}
\usepackage{amsmath}

\title{Extracting Fine-Grained Knowledge Graphs of Scientific Claims:\\ Dataset and Transformer-Based Results}

\author{Ian H. Magnusson$^{\spadesuit\heartsuit}$ \and Scott E. Friedman$^{\spadesuit}$ \\
  $^\spadesuit$SIFT, Minneapolis, MN, USA \\
  $^\heartsuit$Northeastern University, Boston, MA, USA \\
  \texttt{magnusson.i@northeastern.edu} \\
  \texttt{friedman@sift.net} \\}

\begin{document}
\maketitle
\begin{abstract}
Recent transformer-based approaches demonstrate promising results on relational scientific information extraction. Existing datasets focus on high-level description of how research is carried out. Instead we focus on the subtleties of how experimental associations are presented by building \sciclaim, a dataset of scientific claims drawn from Social and Behavior Science (SBS), PubMed, and CORD-19 papers. Our novel graph annotation schema incorporates not only coarse-grained entity spans as nodes and relations as edges between them, but also fine-grained attributes that modify entities and their relations, for a total of 12,738 labels in the corpus. By including more label types and more than twice the label density of previous datasets, \sciclaim captures causal, comparative, predictive, statistical, and proportional associations over experimental variables along with their qualifications, subtypes, and evidence. We extend work in transformer-based joint entity and relation extraction to effectively infer our schema, showing the promise of fine-grained knowledge graphs in scientific claims and beyond.
\end{abstract}


\section{Introduction}

Using relations as edges to connect nodes consisting of extracted entity mention spans produces expressive and unambiguous knowledge graphs from unstructured text. This approach has been applied to diverse domains from moral reasoning in social media \cite{tybalt2021cogsci} to qualitative structure in ethnographic texts \cite{friedman2021qr}, and is particularly useful for reasoning about scientific claims, where several experimental variables in a sentence may have differing relations. Scientific information extraction datasets such as \scierc \cite{luan2018multi} use relations for labeling general scientific language. Utilizing the advances of \scibert \cite{beltagy2019scibert} in scientific language modeling, \spert \cite{eberts2019span}---a transformer-based joint entity and relation extraction model---advanced the state of the art on \scierc.
\footnotetext[1]{Dataset available at https://github.com/siftech/SciClaim.}

\begin{figure}[t]
\centering
\includegraphics[width=\linewidth]{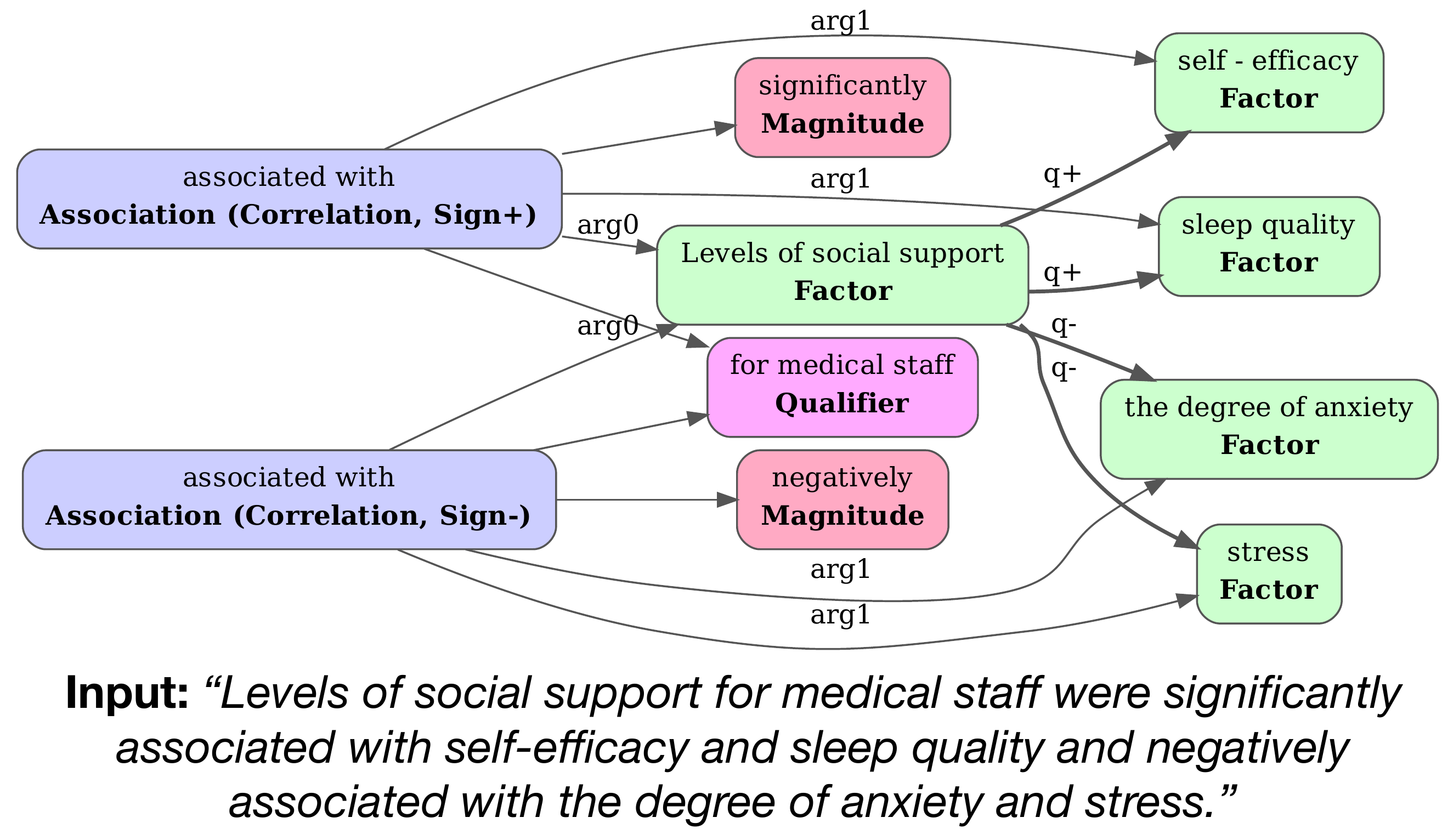}
\caption{\sciclaim knowledge graph with entities (nodes), relations (edges), and attributes (parentheticals) connecting an independent variable via \textit{arg0} to distinct correlations with dependent variables via \textit{arg1}.}
\label{fig:claim1}
\vspace{-0.2in}
\end{figure}

To extend relational scientific information extraction to specifically target scientific claims, we annotate \sciclaim,\footnotemark[1] a dataset of 12,738 annotations on 901 sentences from expert identified claims in Social and Behavior Science (SBS) papers \cite{alipourfard2021OSF}, detected causal language in PubMed papers \cite{yu2019EMNLPCausalLanguage}, and claims and causal language heuristically identified from CORD-19 abstracts \cite{wang2020cord19}. 
\begin{figure*}[htb]
\centering
\includegraphics[width=\textwidth]{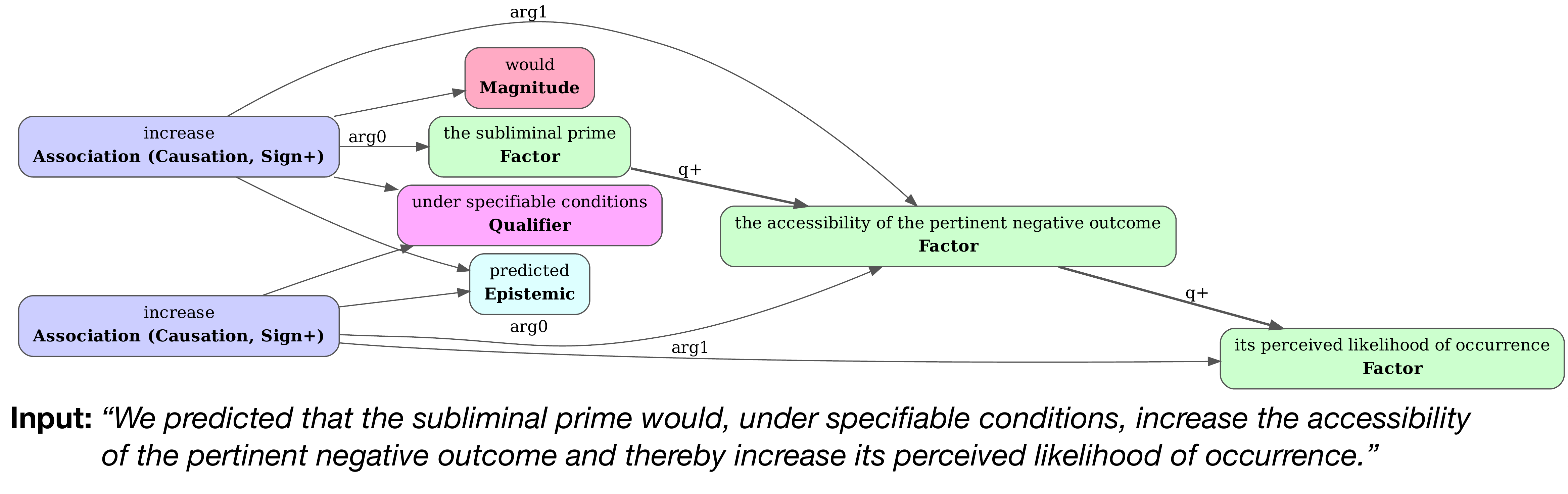}
\caption{This \sciclaim graph captures the chaining together of associations and uncovers a mediating factor in the qualitative proportionality (\textit{q+}) between the "subliminal prime" and "perceived likelihood of occurrence."}
\label{fig:claim2}
\end{figure*}

For annotation, we developed a novel graph schema that reifies claimed associations as entity spans with fine-grained attributes and extracts factors as additional entities connected with relations to one or more associations in which they are involved. In \figref{claim1}, two association entities relate two pairs of dependent factors to an independent factor, while attributes and additional relations delimit the scope and qualitative proportionalities of the claim. Inspired by semantic role labeling, attributes modify associations and the roles of their arguments, allowing us to represent claims of causal, comparative, predictive, statistical, and proportional associations along with their qualifications, subtypes, and evidence. 

We adapt \spert to model this additional multi-label attribute task and demonstrate that extraction of our highly expressive knowledge graphs is within reach of present methods.


\begin{table*}[htb]
\footnotesize
\centering
\begin{tabular}{r ccccccccc }
\toprule
& &  \multicolumn{2}{c}{\textbf{Entities}} & \multicolumn{2}{c}{\textbf{Relations}} & \multicolumn{2}{c}{\textbf{Attributes/Corefs}} & \multicolumn{2}{c}{\textbf{Total Labels}} \\
\cmidrule(lr){3-4}
\cmidrule(lr){5-6}
\cmidrule(lr){7-8}
\cmidrule(lr){9-10}
\textbf{Dataset} & \textbf{Words} & Count & Per Word & Count & Per Word & Count & Per Word & Count & Per Word \\ \midrule 
SciREX      & \textbf{2512806} & \textbf{157680} & 6.27\%       & \textbf{9198} & 0.37\%        & - &   - & \textbf{166878} & 06.64\%\\
SemEval2017 & 84010   & 9946 & 11.84\%       & 672 &  0.79\%        & - & - & 10618 & 12.64\%  \\               
SemEval2018 & 58144   & 7483 & 12.87\%       & 1595 &  2.74\%        & - & - & 9078 & 15.61\%\\                 
\scierc      & 65334   & 8089 & 12.38\%       & 4716 & 7.21\%        & \textbf{2752} & 4.21\% & 15557 & 23.81\% \\            
\sciclaim         & 20070   & 5548 & \textbf{27.64\% }      & 5346 & \textbf{26.64\%}        & 1844 & \textbf{9.19\%} & 12738  &\textbf{63.47\% }

\\\bottomrule
\end{tabular}

\caption{Our \sciclaim dataset contains the highest label densities per word and comparable label counts to other scientific information extraction datasets except SciREX, which uses distant supervision and noisy automatic labeling. Our dataset contains fine-grained attributes as additional labels, while \scierc contains coreference links.}
\label{tab:datasets}
\end{table*}
\section{Related Work}
\label{sec:related_work}
Many previous datasets for relational scientific information extraction—such as SemEval 2017 task 10 and 2018 task 7, SciERC, and SciREX \cite{augenstein-etal-2017-semeval, gabor-etal-2018-semeval, luan2018multi, jain-etal-2020-scirex}—have annotated corpora from NLP, computer science, or similar engineering-oriented fields. As such their annotation schemas have emphasized the description of how research was carried out, by extracting categories of entities such as methods, tasks, metrics, and datasets as well as relations mostly describing their intrinsic properties such as uses, composition, and hyponymy. Two of these datasets \cite{luan2018multi, gabor-etal-2018-semeval} contain associative relations that directly link entities being compared or producing a result. Our work extends further in this direction by examining not only which entities are associated, but also how the presentation of the associations is nuanced by the assertion of fine-grained attributes such as causality or proportionality. 

\sciclaim provides the largest number of fine-grained label types among comparable datasets. \tabref{datasets} shows \sciclaim's remarkable label densities per word. \sciclaim also contains 81.88\% as many total labels as \scierc and more total labels than SemEval 2017 task 10 and 2018 task 7. On the other hand, SciREX utilizes distant supervision from an existing knowledge base and noisy automatic labeling trained on SciERC to provide an order of magnitude more labels and annotate complete documents. This is one example of how smaller yet more densely and directly labeled datasets like \scierc and \sciclaim can enable and compliment larger, higher-level corpora.

 Meanwhile, our dataset also focuses on scientific claims. Some previous work \textit{identifies} claims within scientific texts \cite{wadden2020fact,score2021twosix}, but does not extract the relations and factors within the claims themselves. Other recent symbolic semantic NLP systems do model relational representations of scientific claims (e.g., \citealp{friedman2017learning}), but these approaches rely on rule-based engines with hand tuning, which require NLP experts to maintain and adapt to new domains. Instead we modify \spert \cite{eberts2019span}, a transformer-based method that has been shown to effectively extract relational scientific information on \scierc \cite{luan2018multi}. We extend this model to accommodate our additional multi-label attributes and apply it to our claim graph extraction task.

\section{Approach}

\subsection{Problem Definitions}
\sciclaim defines the multi-attribute knowledge graph extraction task as follows: for a sentence $\mathcal{S}$ of $n$ tokens $s_1,...,s_n$, and sets of entity types $\mathcal{T}_e$, attribute types $\mathcal{T}_a$, and relation types $\mathcal{T}_r$, predict the set of entities $\langle s_j, s_k, t \in \mathcal{T}_e \rangle \in \mathcal{E}$ ranging from tokens $s_j$ to $s_k$, where $1 \leq j \leq k \leq n$, the set of relations over entities $\langle e_{head} \in \mathcal{E}, e_{tail} \in \mathcal{E}, t \in \mathcal{T}_r \rangle \in \mathcal{R}$ where $e_{head} \neq e_{tail}$, and the set of attributes over entities $\langle e \in \mathcal{E}, t \in \mathcal{T}_a \rangle \in \mathcal{A}$.
This defines a directed multi-graph without self-cycles, where each unique span can be represented by at most one entity node with zero to $|\mathcal{T}_a|$ attributes.

\subsection{Dataset Construction}


To create \sciclaim, two NLP researchers annotated 901 sentences from several sources: 336 from papers in Social and Behavior Science (SBS) with expert identified claims \cite{alipourfard2021OSF}, 411 filtered for causal language in PubMed papers \cite{yu2019EMNLPCausalLanguage}, 135 containing claims and causal language identified from CORD-19 abstracts \cite{wang2020cord19} with heuristic keywords, and 19 manual perturbations included only in training data. 

To aid in the labeling of these densely annotated sentences, we iteratively trained on already collected data and utilized the predictions of the partially trained model on new training examples as suggestions in our labeling interface. We disabled these model suggestions on our 100 example test set to ensure that this did not bias our evaluation.

Due to the dense and potentially overlapping span annotations, small decisions about what tokens to include in a span frequently influence the span boundaries of several other entities in a sentence. However, most of these decisions have negligible impact on the meaningfulness of the annotation (e.g. the decision to include a determiner in span), rendering exact match agreement ineffective. 
Instead to promote consistency and domain relevance we employed iterative schema design sessions in consultation with a subject matter expert in reproducibility of SBS experiments and a process of consensus, schema re-development, and re-annotation on 250 examples where annotators overlapped. 


\tabref{datasets} contrasts \sciclaim's label counts and density with the other relational scientific information extraction datasets discussed in Section \ref{sec:related_work}, and precise counts for each label type are provided in \tabref{schema-results}. Further details are in Appendix \ref{sec:appendix_dataset}.

\subsection{Graph Schema}

The \sciclaim graph schema is designed to capture associations between factors (e.g., causation, comparison, prediction, statistics, proportionality), monotonicity constraints across factors, epistemic status, subtypes, and high-level qualifiers. 

\begin{figure}[htb]
\centering
\includegraphics[width=\linewidth]{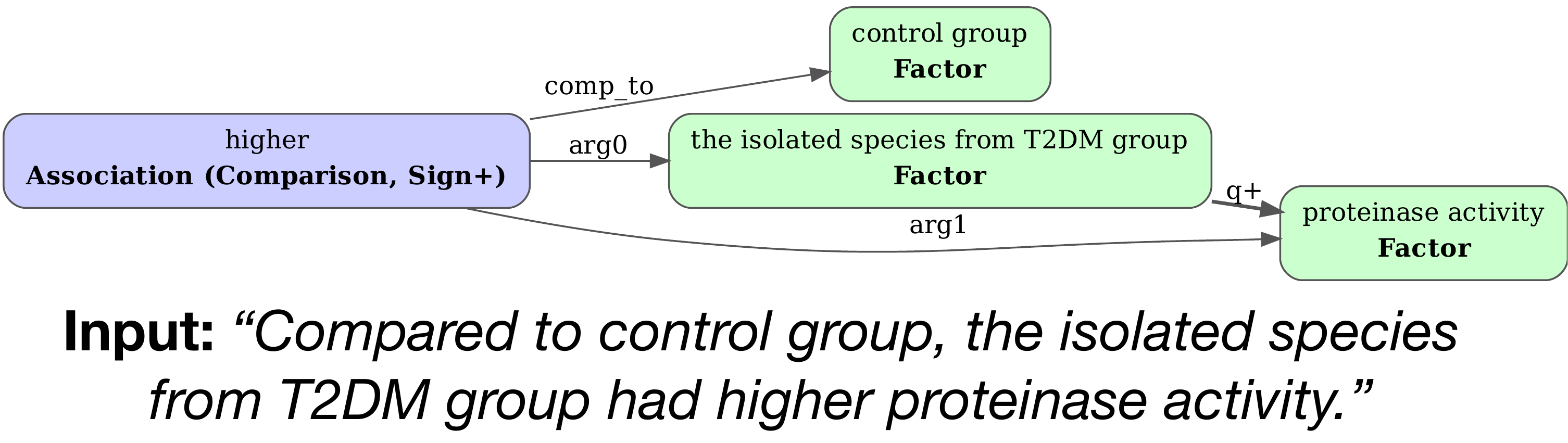}
\caption{\textit{Comparison} attributes modify arguments to account for a (sometimes implicit) frame of reference.}
\label{fig:claim3}
\end{figure}

\textbf{\textit{Entities}} are labeled text spans. The same exact span cannot correspond to more than one entity type, but two entity spans can overlap. Entities comprise the nodes of \sciclaim graphs upon which attributes and relations are asserted.
Our schema includes six entity types:
\textbf{Factors} are variables that are tested or asserted within a claim
 (e.g., \textq{sleep quality} in \figref{claim1}).
\textbf{Associations} are explicit phrases associating one or more factors 
(e.g., \textq{higher} \figref{claim3}).
\textbf{Magnitudes} are modifiers of an association indicating its likelihood, strength, or direction 
(e.g., \textq{significantly} and \textq{negatively} in \figref{claim1}).
\textbf{Evidence} is an explicit mention of a study, theory, or methodology supporting an association 
(e.g., \textq{our SEIR model}).
\textbf{Epistemics} express the belief status of an association, often indicating whether something is hypothesized, assumed, or observed 
(e.g., \textq{predicted} in \figref{claim2}).
\textbf{Qualifiers} constrain the applicability or scope of an assertion 
(e.g., \textq{for medical staff} in \figref{claim1}).

\textbf{\textit{Attributes}} are multi-label fine-grained annotations (visualized in parentheses), where zero or more may apply to any given entity.
Our schema includes the following attributes, all of which apply solely to \textit{Association} entities:
\textbf{Causation} expresses cause-and-effect over its constituent factors 
(e.g., both \textq{increase} spans in \figref{claim2}).
\textbf{Correlation} expresses interdependence over its constituent factors 
(e.g., both \textq{associated with} spans in \figref{claim1}).
\textbf{Comparison} expresses an association with a frame of reference 
(as in the \textq{higher} statement of \figref{claim3}).
\textbf{Sign+} and \textbf{Sign-} expresses high/low or increased/decreased factor value 
(e.g., \textq{correlates more closely with} or \textq{shortened} respectively).
\textbf{Test} expresses statistical measurements 
(e.g., \textq{ANOVA}).
\textbf{Indicates} expresses a predictive relationship 
(e.g., \textq{prognostic factors for}).

\textbf{\textit{Relations}} are directed edges between labeled entities in \sciclaim graphs.
They are critical for expressing what-goes-with-what over the set of entities.
Note that in Figures \ref{fig:claim1} and \ref{fig:claim2} the unlabeled arrows are all \textit{modifier} relations, left blank to avoid clutter.
We encode six relations:
\textbf{arg0} relates an association to its cause, antecedent, subject, or independent variable 
(e.g., \textq{levels of social support} in \figref{claim1}).
\textbf{arg1} relates an association to its result or dependent variable 
(e.g., \textq{self-efficacy} and \textq{stress} in \figref{claim1}).
\textbf{comp\_to} is an explicit frame of reference in a comparative association 
(e.g., \textq{control group} in \figref{claim3}).
\textbf{subtype} relates a head entity to a subtype tail
(e.g., \textq{stillbirth} as a subtype of \textq{pregnancy outcome}).
\textbf{modifier} relates associations to qualifiers, magnitudes, epistemics, and evidence 
(e.g., all unlabeled arrows in \figref{claim1} and \figref{claim2}).
\textbf{q+} and \textbf{q-} indicate positive and negative qualitative proportionality, respectively, where increasing the head factor increases or decreases the tail factor, respectively
(e.g., \textq{levels of social support} is \textit{q+} to \textq{sleep quality} and \textit{q-} to \textq{stress} in \figref{claim1}).

\subsection{Model Architecture}
\label{sec:model_architecture}
\begin{figure}[t]
\centering
\includegraphics[width=\linewidth]{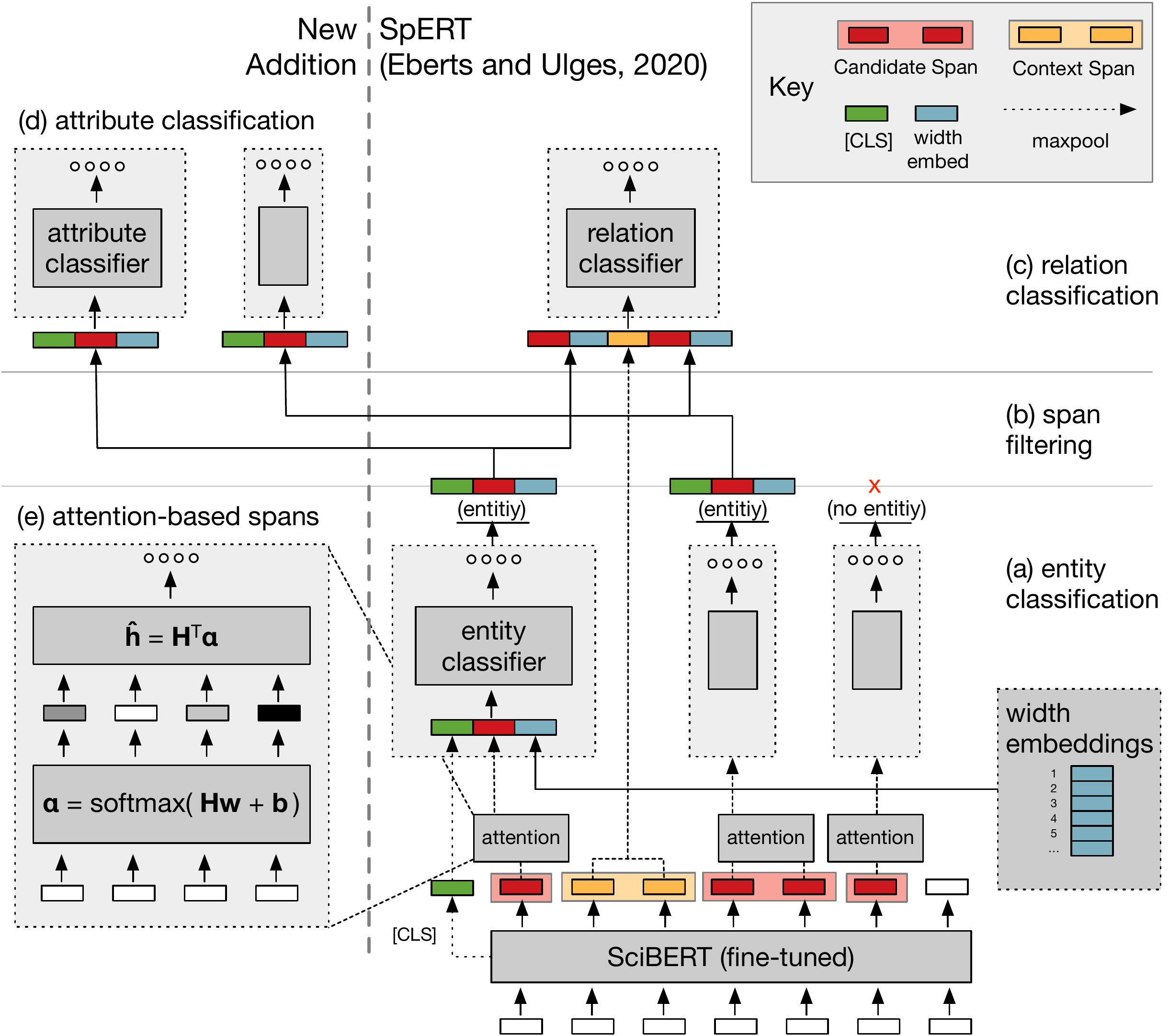}
\caption{Our extension of \spert components (a, b, and c) with multi-label attributes (d) and attention-based entity span representations (e).}
\label{fig:model}
\end{figure}

\begin{table*}[htb]
\footnotesize
\centering
\begin{tabular}{r l ccc ccc ccc}
\toprule
 & & \multicolumn{3}{c}{\textbf{Entities}} & \multicolumn{3}{c}{\textbf{Attributes}} & \multicolumn{3}{c}{\textbf{Relations}} \\
\cmidrule(lr){3-5}
\cmidrule(lr){6-8}
\cmidrule(lr){9-11}
\textbf{Data}& \textbf{Model} & \textbf{P} & \textbf{R} & \textbf{F1} & \textbf{P} & \textbf{R} & \textbf{F1} & \textbf{P} & \textbf{R} & \textbf{F1} \\ \midrule
 \scierc & \spert  & \textbf{70.87} & \textbf{69.79} & \textbf{70.33} & - & - & - & \textbf{53.40} & \textbf{48.54} & \textbf{50.84} \\
\specialrule{.01em}{.3em}{0.3em}


 \sciclaim & \spert-attrs-as-ents & 
 \textbf{90.13} & \textbf{88.63} & \textbf{89.37} & 
 \textbf{92.35} & 82.13 & 86.94 & 
 \textbf{77.59} & \textbf{74.34} & \textbf{75.92} \\

   \sciclaim & \spert-modified & 
 89.81&87.87&88.83&
91.89&\textbf{82.62}&\textbf{87.01}&
76.43&73.72&75.05\\

 \bottomrule
\end{tabular}
\caption{Micro \textbf{P}recision, \textbf{R}ecall, and \textbf{F1} averaged over 5 runs on \sciclaim with \scierc for comparison.}
\label{tab:metrics}
\end{table*}
\begin{table}[htb]
\footnotesize
\centering
\begin{tabular}{c r cccc }
\toprule
& \textbf{Label} & \textbf{P} & \textbf{R} & \textbf{F1} & \textbf{S} \\ \midrule

\parbox{3mm}{\multirow{6}{*}{\rotatebox[origin=c]{90}{\textbf{Entities}}}} 
&factor&91.28&89.97&90.62 & 2756\\
&evidence&88.80&93.33&90.96 & 230\\
&epistemic&91.21&72.17&80.52 & 299\\
&association&92.45&88.20&90.27 & 1290\\
&magnitude&87.71&88.38&88.02 & 613\\
&qualifier&75.86&78.33&77.02 & 360\\
\midrule
\parbox{3mm}{\multirow{6}{*}{\rotatebox[origin=c]{90}{\textbf{Attributes}}}} 
&causation&38.15&60.00&46.20 & 342\\
&comparison&86.69&80.00&83.19 & 329\\
&indicates&84.79&76.25&80.20 & 84\\
&sign+&97.27&88.31&92.58 & 542\\
&sign-&91.97&72.86&81.28 & 202\\
&correlation&98.42&84.41&90.88 & 320\\
\midrule
\parbox{3mm}{\multirow{7}{*}{\rotatebox[origin=c]{90}{\textbf{Relations}}}} 
&arg0&79.53&75.03&77.19 & 1325\\
&arg1&79.92&77.57&78.71 & 1384\\
&comp\_to&65.86&60.00&62.78 & 187\\
&modifier&77.71&76.21&76.95 & 1582\\
&subtype&40.00&33.33&36.00 & 156\\
&q+&65.53&67.61&66.50 & 504\\
&q-&70.70&53.00&60.09 & 208\\

\bottomrule
\end{tabular}
\caption{High \textbf{P}recision, \textbf{R}ecall, and \textbf{F1} across most types relative to total \textbf{S}upport in \sciclaim, using \spert-modified averaged over 5 runs.}
\label{tab:schema-results}
\end{table}

In order to model the additional multi-label task in \sciclaim, we extend \spert \cite{eberts2019span} with an attribute classifier. \spert provides components (\figref{model} a--c) for joint entity and relation extraction and permits the overlapping spans in our data. These classifiers utilize a span representation that combines the \scibert \cite{beltagy2019scibert} contextual embeddings of all tokens in the span through maxpooling, along with a context representation and learned width embedding. \spert classifies entities first and only infers relations on pairs of identified entities. 

Instead of maxpool we adopt an attention-based span representation (\figref{model} e) inspired by \citet{lee-etal-2017-end}. 
This produces scalars $\alpha_{i,t}$ for each SciBERT token vector $\mathbf{h}_t$ in a span $i$ using learned parameters $\mathbf{w}$ and $b$:
\begin{align}
    \alpha_{i,t} = { \exp(\mathbf{w} \cdot \mathbf{h}_t + b) \over \sum_{k=START(1)}^{END(i)} \exp(\mathbf{w} \cdot \mathbf{h}_k + b)}
\end{align}

These attention weights are used to make a span representation $\mathbf{\hat{h}}_i$ with the following weighted sum:
\begin{align}
    \mathbf{\hat{h}}_i = \sum_{t=START(1)}^{END(i)}\alpha_{i,t} \mathbf{h}_t
\end{align}

We use the same cascaded inference strategy and input the span representations of identified entities $\mathbf{x}^a$ to an attribute classifier (\figref{model} d) with weights $\mathbf{W}^a$ and bias $\mathbf{b}^a$. A pointwise sigmoid $\sigma$ yields seperate confidence scores $\mathbf{\hat{y}}^a$ for each attribute:
\begin{align}
    \mathbf{\hat{y}}^a = \sigma(\mathbf{W}^a \mathbf{x}^a + \mathbf{b}^a)
\end{align}
We train the attribute classifier with a binary cross entropy loss $\mathcal{L}_a$ 
summed with the \spert entity and relation losses, $\mathcal{L}_e$ and $\mathcal{L}_r$, for a joint loss:
\begin{align}
   \mathcal{L} = \mathcal{L}_e + \mathcal{L}_r + \mathcal{L}_a 
\end{align}

\section{Evaluation}

In \tabref{metrics} we report micro performance metrics on the \sciclaim test set averaged over 5 runs.

In addition to the \textbf{modified} \spert (detailed in Section \ref{sec:model_architecture}), we also test a variant \textbf{attrs-as-ents} where all attribute labels on an entity span are collapsed into a single combined annotation, allowing unmodified \spert to process attributes. 
Precisely, we collapse $\mathcal{T}_e$ entity types with all combinations of ${\mathcal{T}_a}$ attribute types into $\{\mathcal{T}_e \times {{\mathcal{T}_a} \choose k} : 0 \leq k \leq |\mathcal{T}_a| \}$ multi-class entity labels. 
We hypothesized that the combinatorially larger number of labels required by \textit{attrs-as-ents} would lower performance on rarely occurring combinations. Surprisingly the variants get almost identical results, suggesting that—at least for our data—a single layer classifier can infer the attributes of a span simultaneously just as well as doing so independently.
We tested other model variants that also produced changes $\sim$1\% F1 and thus are relegated to Appendix \ref{sec:variants}.

To our knowledge no previous models exists that can run directly on all three tasks in our dataset due to the presence of both overlapped entity spans and multi-label attributes. For comparison we include \spert's state-of-the-art performance on \scierc, the dataset closest to ours in terms of label density. The high performance of our adapted \spert on \sciclaim demonstrates the practicality of extracting our novel graph schema with present methods despite its fine-grained approach.

The per-class evaluations for our main model are reported in \tabref{schema-results}. With few exceptions performance is good, and generally follows support for the label in the dataset. The \emph{Causation} attribute metrics may be influenced by noise from anomalously low representation in the test set (only 5 instances compared to 59 instances of \emph{Correlation}). Likewise the \emph{Test} attribute unfortunately does not appear in the test set at all, but receives validation F1 of 95.95\% despite only appearing 25 times in the corpus. Another outlier, the \emph{subtype} relation, is particularly challenging, especially with its low rate of occurrence, due to it being one of the few relation types occurring directly between factors rather than mediated through a reified association span. The \emph{q+}/\emph{q-} relations are likewise expressed as direct links between factors. Although these require complex reasoning about the qualitative proportionalities of factors (e.g., \figref{claim2}), they nonetheless receive promising results. The attributes \emph{Sign+}/\emph{Sign-} serve a similar role and provide partial redundancy for \emph{q+}/\emph{q-} labels, allowing downstream reasoning to back off to these less precise, more robust attributes when the qualitative proportionalities are not extracted.

\section{Conclusion}

Previous scientific information extraction crafts useful high-level representation of papers, going as far as document level relations spanning thousands of words in \citet{jain-etal-2020-scirex}. Complimentary to these efforts, we propose fine-grained and densely annotated scientific information extraction that captures not just what is said but how it is presented and argued. \sciclaim applies this approach to associative claims and demonstrates that existing models such as \spert \cite{eberts2019span} can be modified to accurately extract fine-grained knowledge graphs ripe for downstream reasoning.

\section*{Acknowledgements}
This material is based upon work supported by the Defense Advanced Research Projects Agency (DARPA) and Army Research Office (ARO) under Contract No. W911NF-20-C-0002. Any opinions, findings and conclusions or recommendations expressed in this material are those of the author(s) and do not necessarily reflect the views of the Defense Advanced Research Projects Agency (DARPA) and Army Research Office (ARO). We thank the reviewers for their helpful feedback.

\bibliography{main}
\bibliographystyle{acl_natbib}

\appendix

\begin{table*}[ht]

\footnotesize
\centering
\begin{tabular}{r c ccc ccc ccc}
\toprule
 & & \multicolumn{3}{c}{\textbf{Entities}} & \multicolumn{3}{c}{\textbf{Attributes}} & \multicolumn{3}{c}{\textbf{Relations}} \\
\cmidrule(lr){3-5}
\cmidrule(lr){6-8}
\cmidrule(lr){9-11}
\textbf{Model}& \textbf{Avg Val F1} & \textbf{P} & \textbf{R} & \textbf{F1} & \textbf{P} & \textbf{R} & \textbf{F1} & \textbf{P} & \textbf{R} & \textbf{F1} \\ \midrule

\spert-attrs-as-ents & 80.45 &
 90.13 & \textbf{88.63} & 89.37 & 
 \textbf{92.35} & 82.13 & 86.94 & 
 \textbf{77.59} & 74.34 & 75.92 \\

\spert-modified & \textbf{80.66} & 
89.81&87.87&88.83&
91.89&\textbf{82.62}&\textbf{87.01}&
76.43&73.72&75.05\\

\spert-modified-maxpool & 80.22 &
\textbf{90.32}&88.54&\textbf{89.42}&
92.00&80.90&86.09&
76.11&\textbf{75.92}&\textbf{75.99}\\

\spert-modified-unfiltered & 79.99 & 
89.28&88.03&88.64&
91.65&80.74&85.84&
75.62&73.98&74.78\\

 \bottomrule
\end{tabular}
\caption{Micro \textbf{P}recision, \textbf{R}ecall, and \textbf{F1} averaged over 5 runs on the \sciclaim test set as well as F1 averaged over the 3 tasks on 5 runs of \sciclaim validation data (\textbf{Avg Val F1}).}
\label{tab:variants}
\end{table*}

\section{Claims Dataset}
\label{sec:appendix_dataset}

Our English language dataset \sciclaim consists of 901 examples sentences divided into a training set of 721 sentences, a validation set of 80 sentences, and a test set of 100 sentences. The training and validation data contain examples that were labeled from corrected suggestions from a partially trained model, while the test set was labeled from scratch without any model suggestions as a starting point.

Our data from CORD-19 \cite{wang2020cord19} is sampled with the following keywords as a heuristic identification of claims and causal language similar to our expert identified data from PubMed and Social and Behavioral Science (SBS) papers:
associated with, reduce, increase, leads to, led to, our result, greater, less, more, cause, demonstrate, show, improve. 

200 of our sentences (50 from PubMed and 150 from SBS) were selected to intentionally minimize the likelihood of claims and causal language. This includes sentences that discuss factors and other entities present in our schema but either do not contain associations or frame associations in unusual ways such as rhetorical questions. We intend for this data to encourage robustness that maintains correct labels for partial graph extractions rather than simply hallucinating associations in all sentences. We employ the following heuristics to identify this data: We sample 50 PubMed sentences from \citet{yu2019EMNLPCausalLanguage} that are identified as having low causal content. We sample 100 titles from SBS paper present in \citet{alipourfard2021OSF}, as titles contain factors but rarely contain explicit associations and may be present in input data from automatically extracted text from PDFs. Finally we sample 50 first lines of SBS papers from \citet{alipourfard2021OSF}, as these lines frequently introduce topics or rhetorical questions which either lack associations or present highly hypothetical associations unlike those in our main corpus.

Each filtered data source was sampled chronologically.

We utilized the following procedure for labeling: The annotators undertook extensive, iterative schema design sessions in consultation with a subject matter expert in reproducibility of SBS experiments. After the schema was settled on pilot examples, a lead annotator established the annotation standards on several hundred examples through a process of relabeling and retraining the suggestion model. Once the suggestion model became effective, the lead annotator and model suggestions guided the other annotator in adopting the annotation standards. The lead annotator reviewed and corrected the 250 overlapping examples in a consensus process with the other annotator.

\section{Variants and Hyperparameters}
\label{sec:variants}
\subsection{Variants}

We experiment with several variants none of which substantially outperformed the others. \textbf{\spert-modified-maxpool} contains our modifications but simply uses \spert's original maxpooling span representation instead of the attention-based representations inspired by \citet{lee-etal-2017-end}. \textbf{\spert-modified-unfiltered} forgoes cascading inferences and instead classifies all possible spans for attributes. Full test and averaged validation results are presented in Table \ref{tab:variants}. 

\subsection{Hyperparameters}
The following hyperparameters and settings were selected using manual tuning of 10-fold cross validation on the training set and optimizing for average micro-f1 performance over all 3 tasks:
language model \scibert uncased,
mini batch size 8,
epochs 40,
optimizer AdamW,
linear scheduling,
warm up 0.05,
learning rate 5e-5,
learning rate warm up 0.1,
weight decay 0.01,
max grad norm 1.0,
size embedding dimension 25,
dropout probability 0.1,
maximum span size 20,
attribute filter threshold 0.55,
relation filter threshold 0.4.

We ran 32 trials on 5 RTX 2080 ti GPUs, where each trial takes roughly 20 minutes. Our model contains 110 million parameters.

We explored the following hyperparameter bounds:
language model $\in$ \{BERT, \scibert, SpanBERT, \scibert tuned on \scierc\},
epochs $\in$ \{20, 40, 80\},
batch size $\in$ \{4, 8, 16\},
learning rate $\in$ \{1e-5, 5e-5, 1e-4\},
scheduling $\in$ \{linear, cyclic \},
warm up $\in$ \{0.0, 0.05, 0.1, 0.15. 0.2, 0.25, 0.3\},
attribute filter threshold $\in$ \{0.4, 0.5, 0.55, 0.6\},
relation filter threshold $\in$ \{0.35, 0.4, 0.5, 0.6\}.
The remaining settings we inherit from \spert as initial experimentation on early datasets revealed little impact.

\end{document}